\documentclass[runningheads]{llncs}
\usepackage[T1]{fontenc}
\usepackage{graphicx}
\usepackage{amsmath}
\usepackage{makecell}
\usepackage{fontawesome}
\usepackage{bigfoot}
\DeclareNewFootnote{Free}[fnsymbol]
\usepackage{color}
\usepackage[bookmarks=false, colorlinks, allcolors=blue]{hyperref}

\usepackage[capitalize]{cleveref}
\crefname{section}{Sect.}{Sects.}
\Crefname{section}{Section}{Sections}
\Crefname{table}{Table}{Tables}
\begin{document}
\title{Feature Fusion Based on Mutual-Cross-Attention Mechanism for EEG Emotion Recognition}
%
\titlerunning{Feature Fusion Based on MCA Mechanism for EEG Emotion Recognition}
%
\author{Yimin Zhao\orcidID{0009-0007-9615-0386} \and
Jin Gu\textsuperscript{*(\faEnvelopeO)}\orcidID{0000-0003-1147-6170}}
%
\authorrunning{Y. Zhao et al.}
%
\institute{School of Computing and Artificial Intelligence, Southwest Jiaotong University, Chengdu, China \\
\email{gujin@swjtu.edu.cn}}
\maketitle             
\begin{abstract}
An objective and accurate emotion diagnostic reference is vital to psychologists, especially when dealing with patients who are difficult to communicate with for pathological reasons. Nevertheless, current systems based on Electroencephalography (EEG) data utilized for sentiment discrimination have some problems, including excessive model complexity, mediocre accuracy, and limited interpretability. Consequently, we propose a novel and effective feature fusion mechanism named Mutual-Cross-Attention (MCA). Combining with a specially customized 3D Convolutional Neural Network (3D-CNN), this purely mathematical mechanism adeptly discovers the complementary relationship between time-domain and frequency-domain features in EEG data. Furthermore, the new designed Channel-PSD-DE 3D feature also contributes to the high performance. The proposed method eventually achieves \textbf{99.49\% (valence)} and \textbf{99.30\% (arousal)} accuracy on DEAP dataset. \footnoteFree{Corresponding author.}

\end{abstract}

\section{Introduction}
Autism and depression are serious psychological problem, potentially leading to detrimental outcomes. A recent study indicated that dysarthria, mood disorders, rumination, literal understanding of problems or communication difficulties make their assessment difficult~\cite{Hervas2023}. Therefore, it is essential for psychological therapist to examine more reliable indicators such as EEG data from patients. By promptly integrating the emotion judgements derived from these signals into the diagnostic process, psychologists are better equipped to formulate tailored treatment strategies for their patients. 

In recent years, the academic community has achieved some advances in emotion recognition through various methods~\cite{Gao2022,Salama2018a,Wang2022}. Initially, the focus was on singular traditional EEG features, such as Differential Entropy (DE)~\cite{Yang2020} and Power Spectral Density (PSD)~\cite{Duan2012}. Subsequently, approaches involving feature fusion and deep learning were adopted to enhance recognition accuracy. With the application of these new technologies, the performance of the classifier has been improved, but there are still some problems. Currently, the mainstream fusion methods implemented different learnable models to extract feature mappings, and then concatenated them directly~\cite{Liu2018,Liu2021c,Jia2020,Sun2022}. Some projects~\cite{Wang2022,Gao2022} appended extra neural networks to further process the feature mappings.
These strategies, which train the networks to autonomously concentrate on significant aspects of the signal, escalates the burden of model training and diminishes the efficiency of the model's outputs. Considering that sentiment classification systems require instantaneous output in practical applications, current increasingly complicated neural networks are not beneficial. In addition, this is not conducive to the interpretability of the task, potentially resulting in moral hazards.

Furthermore, the latest study~\cite{Li2023} indicated an emerging trend of utilizing 3D data inputs for models. The review identified two predominant structures of Channel-Time-Frame~\cite{Salama2018a} and Channel-Topology-Time~\cite{Wang2018}. However, the final results were unsatisfactory. As shown in \cref{tab:other-3D-SOTA}, the accuracy of 2D-Topology-DE structure of Yang et al.~\cite{Yang2018} is only 90.24\%, which could be the SOTA 3D input feature structure of the other projects that use 3D-CNN network models. Our analysis suggests that the limited spatial information provided by the channel topology map may contribute to this situation.

In that case, to achieve an instant well-performing emotion justification system based on EEG analysis, this project introduces a novel solution that has been experimentally validated as the new state-of-the-art (SOTA) method. It encompasses two primary contributions:

\subsubsection{Mutual-Cross-Attention Mechanism.}
Inspired by the self-attention mechanism proposed by Vaswani et al.~\cite{Vaswani2017}, we introduce a purely mathematical method named Mutual-Cross-Attention (MCA) for it applies Attention Mechanism from each directions of two features. In the field of EEG emotion analysis, we are the first to propose a pure mathematical fusion method, coupled with customized 3D-CNN, to accomplish the task of feature fusion.

\subsubsection{New 3D feature presentation.}
By analyzing existing projects, it is found that spectral information might be more prominent than spacial information (presented by channel topology). Hence, we develop a unique Channel-Frequency-Time 3D feature structure. This innovative feature presents spectral and temporal information simultaneously.


\section{Methods}
To evaluate the proposed MCA, we designed a complete experimental pipeline with five steps: Data Acquisition, Pre-process, Feature Extraction, Feature Fusion, and Classification. In terms of the feature fusing procedure, the complementarity between multiple features and the ability of the fusion mechanism to find important information are both crucial. Finding the optimal combination is challenging. It is widely recognized that DE and PSD complement each other~\cite{Luck2014}. Hence, these two features are selected for further feature fusion research. The accuracy results, based on the Circumplex Model of Affect that concentrates on arousal and valence, are compared with other SOTA methods to demonstrate the validity of our methodology

\subsection{Data acquisition}
The DEAP dataset from Queen Mary University was selected for our experimental setup. In the study, 32 individuals were monitored using electroencephalogram (EEG) and peripheral physiological signals as they viewed 40 one-minute music video clips. Participants rated each video on a scale of 1 to 9 in terms of arousal, valence, likeability, dominance, and familiarity~\cite{Koelstra2012}. The data acquisition equipment has 32 channels and work with 512 Hz sampling frequency.
\subsection{Pre-process}
The dataset is pre-processed guided by the well-known steps of Steve Luck~\cite{Luck2014}. It is cleaned by filtering wave and excluding noise components.
Firstly, Notch Filter is implemented to eliminate 50 Hz signal, commonly associated with interference from AC power sources. Additionally, considering measurement tool inaccuracies and environmental interferences, a 4-45 Hz band-pass filter is set.
Following this, Independent Component Analysis (ICA) is applied to the filtered EEG to cancel noise elements like Electrooculogram, Electrocardiogram, Electromyography.
The final step involved downsampling the original 512 Hz data to 128 Hz. These operations improve data quality, reduce data volume, and accelerate computation speed.


\subsection{Feature extraction}
\label{sec:feature-extraction}
The DE and PSD extractions are adopted across five distinct frequency bands to enhance feature representation and prevent information from influencing each other. The categories include $\theta$ (4-7 Hz), $\alpha$ (8-10 Hz), slow $\alpha$ (8-13 Hz), $\beta$ (14-29 Hz), and $\gamma$ (30-45 Hz).

\subsubsection{DE extraction.}
There are several methods to calculate DE. If the signal fits the Gaussian distribution, which is performed as $N(\mu, \epsilon^2)$. The mathematical formulation is equal to the following one:

\begin{align}
  h(X)=& -\int_{-\infty}^{\infty}{\frac{1}{\sqrt{2\pi\epsilon^2}}e^{-\frac{(x-\mu)^2}{2\epsilon^2}}\log{\left(\frac{1}{\sqrt{2\pi\epsilon^2}}e^{-\frac{(x-\mu)^2}{2\epsilon^2}}\right)}dx} \notag\\
  =& \frac{1}{2}log\left(2\pi e\epsilon^2\right)
  \label{eq:DE-2}
\end{align}

where $\epsilon$ is the standard deviation of $f(x)$. It has been proven that the EEG data filtered by a 4-45 or similar band-pass filter fits a Gaussian distribution every 2 Hz~\cite{Shi2013}. The formulation is applied every 2 seconds of data. Then these segments of DEs are collected and constructed as a DE trial array. Finally, the band-pass filter is applied to consider the DE feature separately according to that 5 different frequency bands mentioned in \cref{sec:feature-extraction}.

\subsubsection{PSD extraction.}
The Welch’s method is used to extract power spectra density (PSD). The first step to acquiring the PSD value is dividing the whole signal into $K$ batches and calculating for each of them. The mathematical presentation of the $k_{th}$ PSD value on frequency $f$ is~\cite{Solomon1991}:

\begin{equation}
  p_k(f)=\frac{1}{W}|F_k(f)|^2
  \label{eq:PSD-1}
\end{equation}

where $W$ is related to the Hanning window and $F_k(f)$ is a windowed fast Fourier transform (FFT) at a specific frequency $f$, which is set as 128 Hz according to the analysis above. The window size is 2 seconds. Finally, the estimation of PSD with the Welch method is combined with the results from all segments:

\begin{equation}
  P_s(f)=\frac{1}{K}\sum_{k=1}^{K}{p_k(f)}
  \label{eq:PSD-3}
\end{equation}

\begin{figure}[t]
\centering
\includegraphics[width=0.95\textwidth]{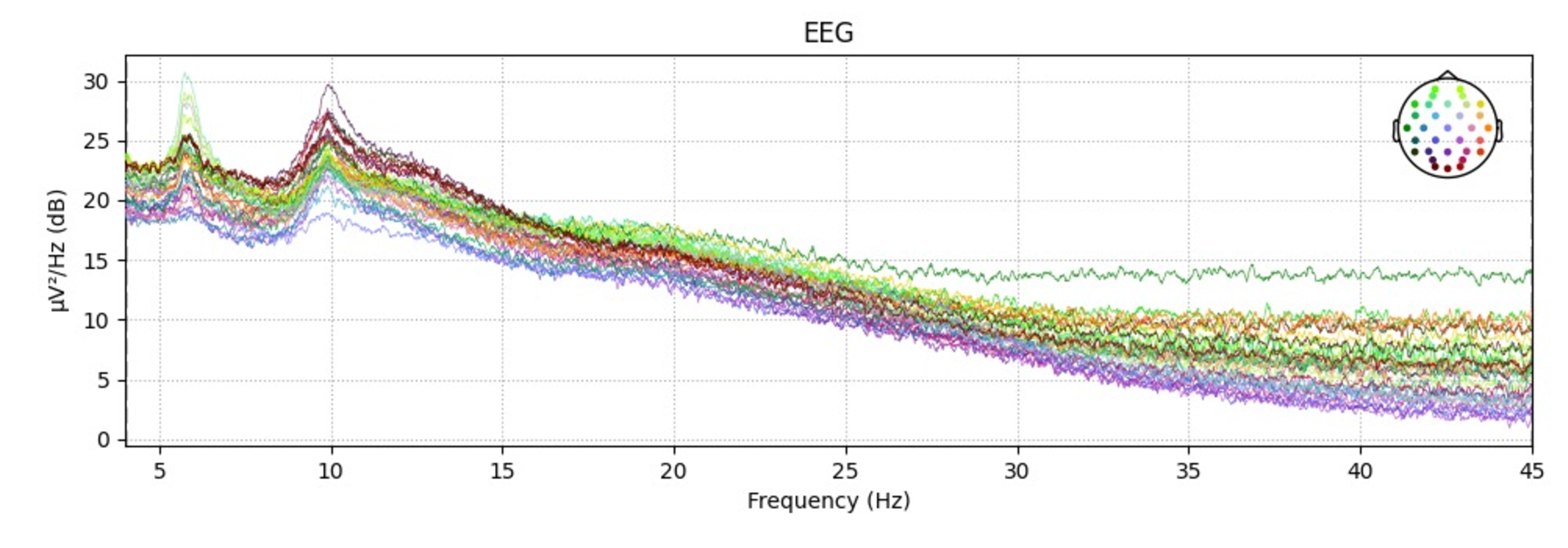}
\caption{PSD diagram of subject 01. } 
\label{fig:PSD}
\end{figure}

To preliminarily evaluate the validity of the PSD, a corresponding diagram is plotted. \cref{fig:PSD} indicates that the local value of the PSD fluctuates in the frequency range of about 5-7 Hz and 9-11 Hz, which suggests there might be emotion presentation in these ranges. That is the reason for the \cref{sec:feature-extraction} indicating two frequency bands (slow $\alpha$ and $\alpha$) in 8-13 Hz range. Finally, the PSD's 4-45 Hz spectrum is categorically divided into five bands for further analysis.

\subsection{Feature fusion}
The MCA mechanism is applied across each selected frequency band to fuse DE and PSD. Initially, respectively consider DE and PSD as key and query vector. Then, designate PSD as value and implement basic Scaled Dot-Product Attention, which is presented by:

\begin{equation}
  \text{Atten}(Q, K, V) = \text{softmax}\left(\frac{QK^T}{\sqrt{d_k}}\right)V
  \label{eq:basic-attention}
\end{equation}

where Q, K, V respectively represent query, key, and value. And $d_k$ is the size of the query's last dimension. 

\begin{figure}[t]
\centering
\includegraphics[width=0.9\textwidth]{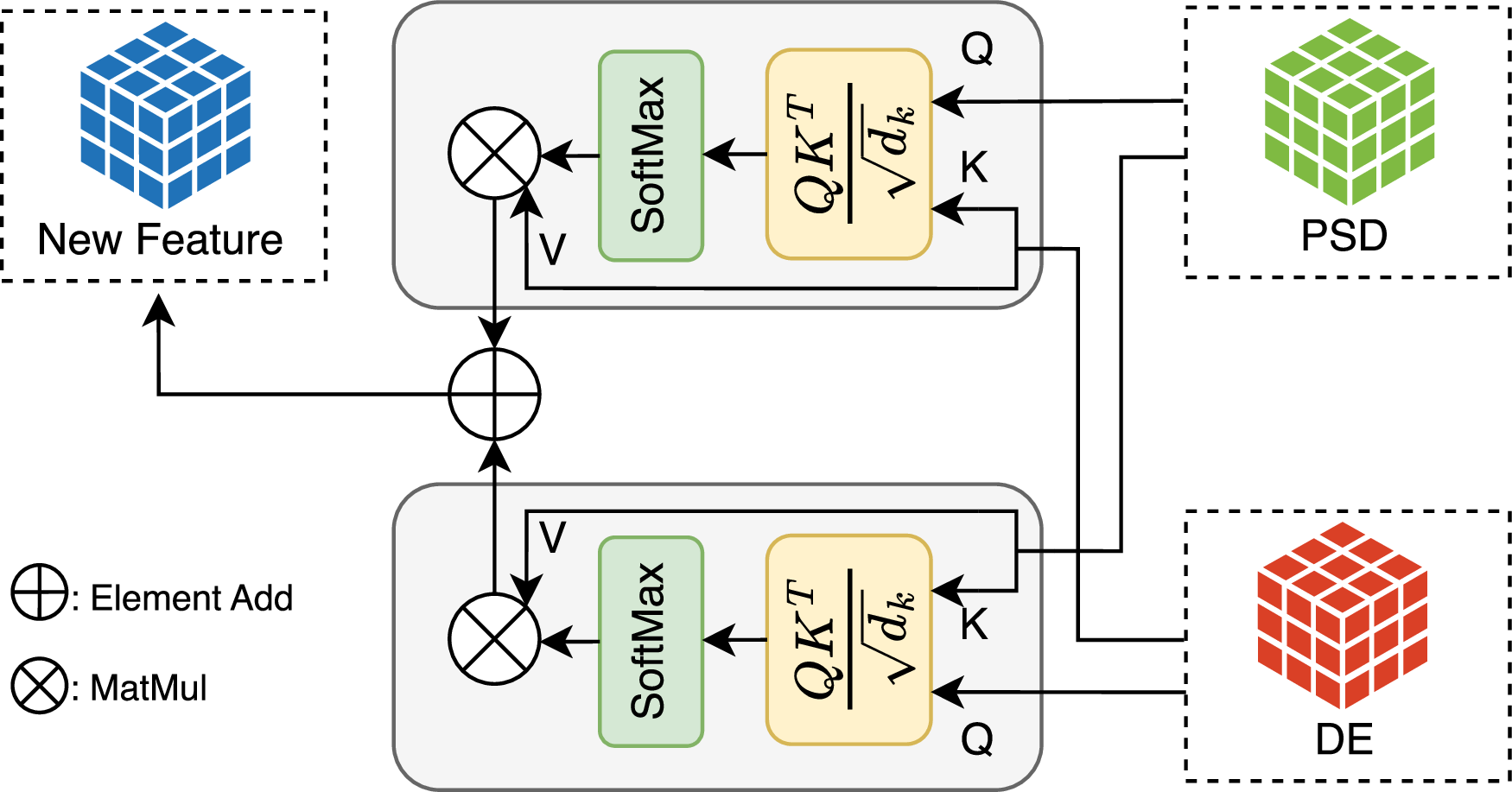}
\caption{Overview of mutual-cross-attention mechanism.} 
\label{fig:MCA}
\end{figure}

That is one direction calculation in MCA. After that, PSD is used as Q, DE as K and V. Again, the Scaled Dot-Product Attention operation is implemented. The results from both directions are added together to get the new feature. \cref{fig:MCA} illustrates the entire process, and its mathematical presentation is:

\begin{equation}
  \text{MCA}(f_1, f_2) = \text{Atten}(f_1, f_2, f_2)+\text{Atten}(f_2, f_1, f_1)
  \label{eq:MCA}
\end{equation}

where $f_1$ is the first feature (DE) and $f_2$ is the second feature (PSD).

\subsection{Classification}
After those feature extraction operations are implemented, the final single feature is denoted as $F_f \in \bbbr^{32 \times 5 \times 60}$. However, it takes too long to perform classification tasks. Therefore, every $F_f$ is split into 20 $F_s \in \bbbr^{32 \times 5 \times 3}$. This operation allows the model to output a sentiment prediction every 3 seconds.

\begin{figure}[t]
\includegraphics[width=\textwidth]{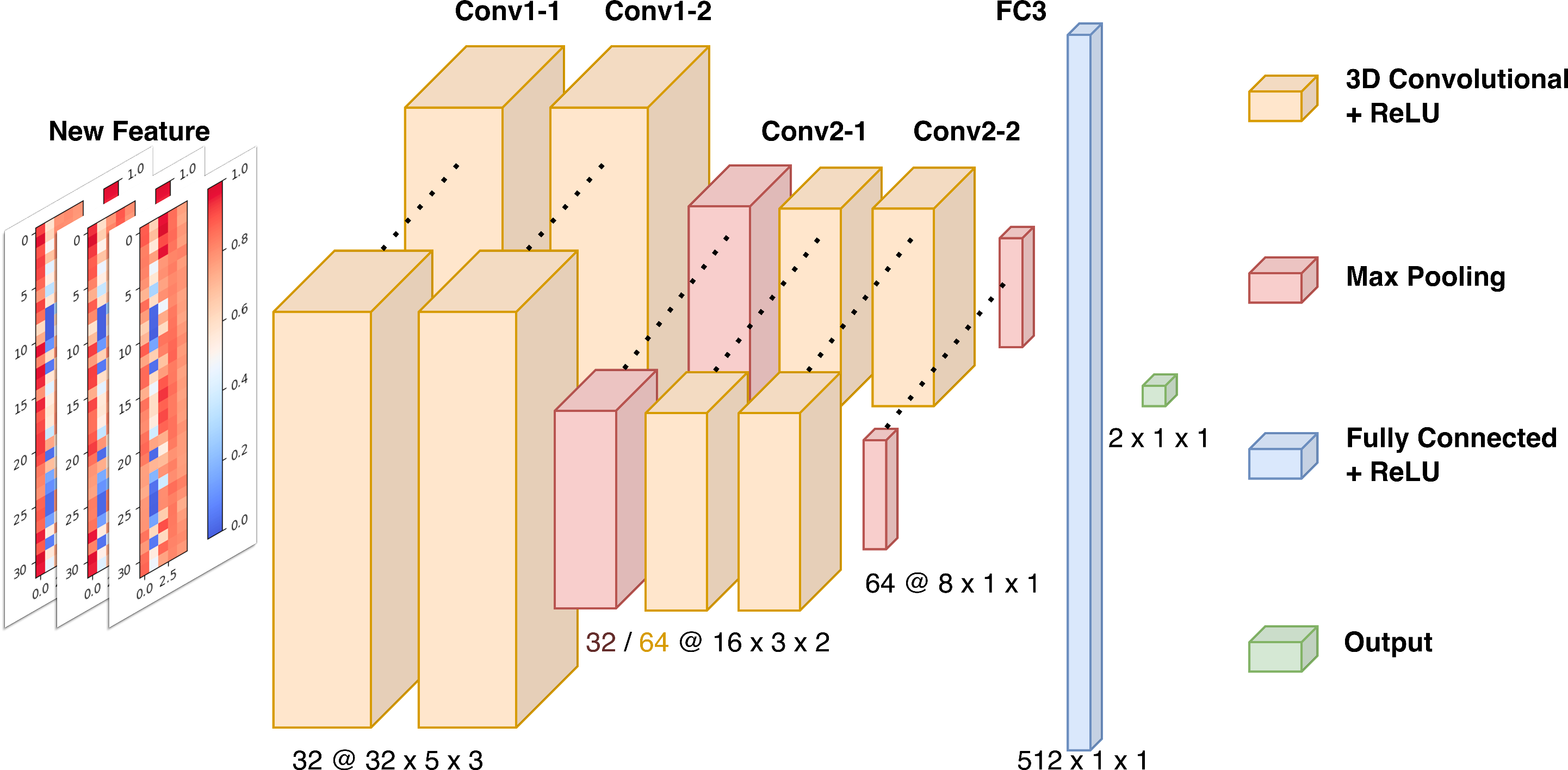}
\caption{Structure of new designed 3D-CNN.} 
\label{fig:3D-CNN-EEG}
\end{figure}

A special 3D-CNN structure is proposed to process the feature $F_s$ . As shown in \cref{fig:3D-CNN-EEG}, the network begins with a 3D convolutional layer, configured with one input channel and 32 output channels, utilizing a 3x3x3 kernel. This layer is followed by another 3D convolutional layer, which maintains the same number of output channels and kernel size. Using two consecutive convolutional layers with the same number of channels enhances the network. The trick deepens the network's capacity to extract features without immediately reducing the spatial dimensions of the input data. After the initial convolutional stage, a 2x2x2 kernel Max Pooling layer is employed with a (0, 1, 1) asymmetric padding.

Subsequently, the network extends into another set of convolutional layers, where the number of output channels is doubled to 64. Following this, the second Max Pooling layer further downsamples the feature maps. Finally, the network transitions to a fully connected layer, which classifies the extracted features into two categories.

\section{Results and discussions}
The performance of the proposed model is demonstrated through various evaluation metrics as detailed in \cref{tab:metrics}. It is clear that all indices exceed 99\%. Additionally, this section includes not only ablation experiments but also comparisons with other SOTA results. All these experiments and comparisons are conducted using the DEAP dataset. Ultimately, the proposed methodology is proven to be effective.

\begin{table}
\centering
\caption{Valence and arousal evaluation metrics of MCA-3D-CNN.}\label{tab:metrics}
\begin{tabular}{l|l|l|l|l}
\hline
Category & Accuracy(\%) & Precision(\%) & Recall(\%) & F1-Score(\%) \\
\Xhline{2\arrayrulewidth}
Valence & 99.49 & 99.60 & 99.54 & 99.57\\
Arousal & 99.30 & 99.45 & 99.31 & 99.38\\
\hline
\end{tabular}
\end{table}

\subsection{Ablation experiments}
The experiment primarily examines the impact of both singular and fused features on the results. For comparison with our proposed MCA method, a baseline configuration labeled "DE+PSD" is established, which is based on the summation of 3D-DE and 3D-PSD. According to the \cref{tab:Ablation}, the accuracy results for single DE and single PSD are almost the same to those of "DE+PSD". However, the valence at 99.49\% and arousal at 99.30\% achieved by the proposed method are significantly higher than those of "DE+PSD". This proves that the proposed method has advantages in the complementary integration of DE and PSD information.

\begin{table}
\centering
\caption{Ablation experiments results. "DE+PSD" represents the element-wise summation between 3D-DE and 3D-PSD.}\label{tab:Ablation}
\begin{tabular}{l|l|l}
\hline
Feature & Valence(\%) & Arousal(\%) \\
\Xhline{2\arrayrulewidth}
Channel-Frequency-DE & 89.88 & 88.16\\
Channel-PSD-Time & 91.88 & 91.56\\
DE+PSD(baseline) & 90.90 & 91.30\\
Proposed MCA & \textbf{99.49} & \textbf{99.30}\\
\hline
\end{tabular}
\end{table}

\subsection{Compare with other SOTA}
The innovation of this project focuses mainly on the design of the new feature structure and the way of fusing the new features. Hence, the comparisons with other SOTA methods in these two aspects are carried out.

\begin{table}
\centering
\caption{Compare with results based on other fusion methods.}\label{tab:other-fusion-SOTA}
  \begin{tabular}{l|l|l|l}
    \hline
    Features & Fusion Method(s) & Valence(\%) & Arousal(\%) \\
    \Xhline{2\arrayrulewidth}
    ResNet, LFCC & Concat \& KNN~\cite{Liu2018} & 90.39 & 89.06 \\ 
    DE, PSD, Hjorth, SE & CNN \& SVM~\cite{Gao2022} & 75.22 & 80.52 \\ 
    PSD, temporal statistics & STFFNN~\cite{Wang2022} & 85.40 & 86.20 \\ 
    Time, 2D-Topology-Time & TSFFN~\cite{Sun2022} & 98.27 & 98.53 \\ 
    \hline
    DE, PSD & MCA \& 3D-CNN~(Ours) & \textbf{99.49} & \textbf{99.30} \\ 
    \hline
  \end{tabular}
\end{table}

\subsubsection{Feature fusion comparison.}
There are various other fusion methods between different features as detailed in \cref{tab:other-fusion-SOTA}. Gao et al.~\cite{Gao2022} integrated DE, PSD, Hjorth, and Sample Entropy (SE). Our project, however, utilizes a narrower range of features. Additionally, similar to the work~\cite{Wang2022}, our project fuses both frequency domain and time domain features. This demonstrates that our proposed method is effective when working with similar features, whether they are the same or fewer in categories.

In the study by Liu et al.~\cite{Liu2018}, the accuracy results are around 90\%, showing commendable performance. However, their approach relies on pre-trained features, which might limit its ability to instantly output results compared to our proposed method. Sun et al.~\cite{Sun2022} developed TSFFN to fuse EEG features with high accuracy. Comprising a 3D-CNN and a transformer, the TSFFN might be too complex for efficient computation. This highlights the advantages of our proposed purely mathematical MCA to feature fusion.

\begin{table}
\centering
\caption{Compare with results based on other singular 3D feature presentations.}\label{tab:other-3D-SOTA}
\begin{tabular}{l|l|l|l}
    \hline
    Feature & Network & Valence(\%) & Arousal(\%) \\
    \Xhline{2\arrayrulewidth}
    2D-Topology-DE~\cite{Yang2018} & 3D-CNN & 89.78 & 90.24 \\ 
    2D-Topology-Time~\cite{Wang2018} & 3D-CNN & 72.10 & 73.10\\ 
    Channel-Time-Frame~\cite{Salama2018a} & 3D-CNN & 87.44 & 88.49 \\ 
    \hline
    Channel-Frequency-DE~(Ours) & 3D-CNN & 89.88 & 88.16 \\
    Channel-PSD-Time~(Ours) & 3D-CNN & \textbf{91.88} & \textbf{91.56} \\
    \hline
\end{tabular}
\end{table}

\subsubsection{Feature structure comparison.}
For 3D feature presentations, the majority of emotion recognition projects based on EEG analysis have inclined towards using topology to expand a 1D channel into a 2D format. Subsequently, data on other dimensions are combined with the topological channel map. As indicated in \cref{tab:other-3D-SOTA}, the performance of our proposed feature structure outperforms all 3D-CNN methods that employ features of the topology and the Channel-Time-Frame. This validates the rationality and effectiveness of the structure we have designed.

\section{Conclusion}
The proposed MCA mechanism, 3D feature Channel-PSD-DE, and customized 3D-CNN show excellent capabilities in EEG-based emotion recognition. The whole system effectively overcomes the limitations of existing systems in terms of instantaneity, accuracy, and interpretability. By integrating DE and PSD features through the MCA mechanism, the ability of mathematical fusion methods to extract meaningful information from EEG data is highlighted. The emotional discrimination system developed using this solution has great potential for practical clinical psychotherapy. In the future, we will further explore the transformer implementing MCA as a core, which could fuse features of larger and more complex datasets.

\newpage

%
%
%
\bibliographystyle{splncs04}
\bibliography{mybibliography}
\end{document}